\documentclass[letterpaper]{article} 
\usepackage{aaai25}  
\usepackage{times}  
\usepackage{helvet}  
\usepackage{courier}  
\usepackage[hyphens]{url}  
\usepackage{graphicx} 
\usepackage{natbib}  
\usepackage{caption} 
\usepackage{algorithm}
\usepackage{algorithmic}

\usepackage{multirow}
\usepackage{threeparttable}
\usepackage{subcaption}
\usepackage{enumitem}
\usepackage{ragged2e}

\usepackage{booktabs}

\usepackage{newfloat}
\usepackage{listings}
\DeclareCaptionStyle{ruled}{labelfont=normalfont,labelsep=colon,strut=off} 
\floatstyle{ruled}
\newfloat{listing}{tb}{lst}{}
\floatname{listing}{Listing}
\usepackage{amsmath}
\usepackage{amssymb}
\usepackage{url}

\title{Consistency of Compositional Generalization across Multiple Levels}
\author{
    Chuanhao Li\textsuperscript{\rm 1,2},
    Zhen Li\textsuperscript{\rm 1},
    Chenchen Jing\textsuperscript{\rm 3}$^*$,
    Xiaomeng Fan\textsuperscript{\rm 1},
    Wenbo Ye\textsuperscript{\rm 2,1}\\
    Yuwei Wu\textsuperscript{\rm 1,2}\thanks{Corresponding author: Chenchen Jing and Yuwei Wu},
    Yunde Jia\textsuperscript{\rm 2,1}
}
\affiliations{
    \textsuperscript{\rm 1}Beijing Institute of Technology
    \textsuperscript{\rm 2}Shenzhen MSU-BIT University
    \textsuperscript{\rm 3}Zhejiang University\\
    \{lichuanhao, li.zhen, fanxiaomeng, yewenbo, wuyuwei, jiayunde\}@bit.edu.cn,
    jingchenchen@zju.edu.cn
}

\begin{document}

\maketitle

\begin{abstract}
    Compositional generalization is the capability of a model to understand novel compositions composed of seen concepts.
    There are multiple levels of novel compositions including phrase-phrase level, phrase-word level, and word-word level.
    Existing methods achieve promising compositional generalization,
    but the consistency of compositional generalization across multiple levels of novel compositions remains unexplored.
    The consistency refers to that a model should generalize to a phrase-phrase level novel composition, and phrase-word/word-word level novel compositions that can be derived from it simultaneously.
    In this paper, we propose a meta-learning based framework, for achieving consistent compositional generalization across multiple levels.
    The basic idea is to progressively learn compositions from simple to complex for consistency.
    Specifically, we divide the original training set into multiple validation sets based on compositional complexity,
    and introduce multiple meta-weight-nets to generate sample weights for samples in different validation sets.
    To fit the validation sets in order of increasing compositional complexity,
    we optimize the parameters of each meta-weight-net independently and sequentially in a multilevel optimization manner.
    We build a GQA-CCG dataset to quantitatively evaluate the consistency.
    Experimental results on visual question answering and temporal video grounding, demonstrate the effectiveness of the proposed framework.
    We release GQA-CCG at \url{https://github.com/NeverMoreLCH/CCG}.
\end{abstract}

\section{Introduction}

\begin{figure}[t]
    \centering
    \includegraphics[width=1\linewidth]{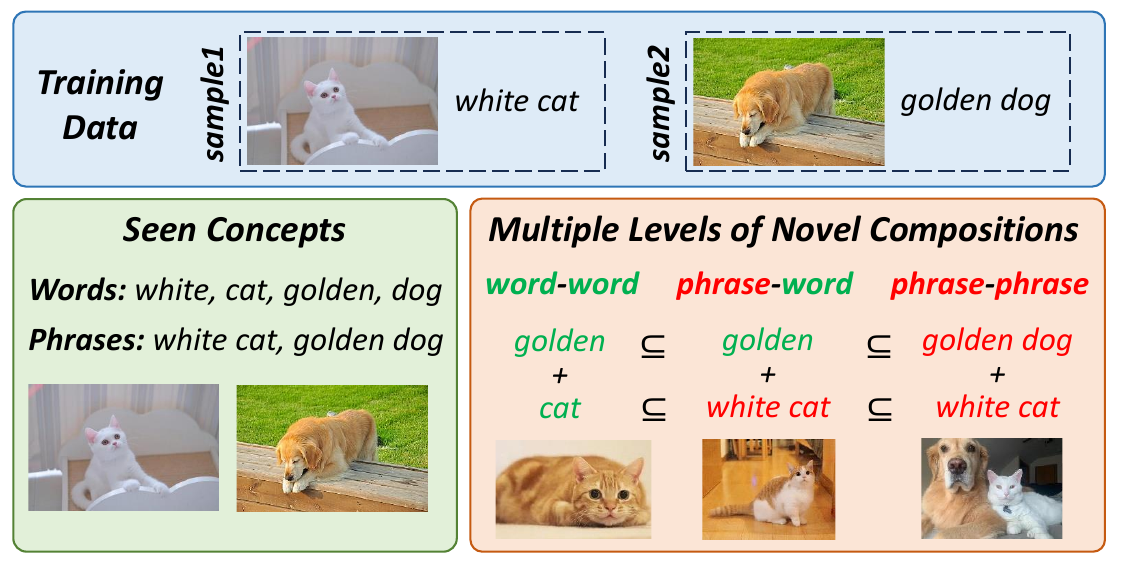}
	\caption{
             Illustration of multiple levels of novel compositions, including word-word level, phrase-word level and phrase-phrase level.
             }
    \label{fig:motivation}
\end{figure}

Compositionality is an important property of human cognition \cite{cognition2}.
Compositional generalization, the capability of a model to understand novel compositions composed of seen concepts, is critical for artificial intelligence systems to mimic the compositionality.
Previous work \cite{pierrot2019learning,liu2020compositional,yang2023deco,xu2023meta} has shown that novel compositions exist at multiple levels, including phrase-phrase level, phrase-word level and word-word level, as shown in Figure \ref{fig:motivation}, but the consistency of compositional generalization across multiple levels of novel compositions remains unexplored.
The consistency refers to the model's ability to generalize to both phrase-phrase level novel compositions and phrase-word/word-word level novel compositions, which are derived from the words within the phrase-phrase structures.
For example, when a model generalize to a phrase-phrase level composition like \textit{``golden dog''}+\textit{``white cat''}, it should also be able to generalize to phrase-word and word-word level compositions, such as \textit{``golden''}+\textit{``white cat''} and \textit{``golden''}+\textit{``cat''}.
Understanding \textit{``golden''}+\textit{``white cat''} and \textit{``golden''}+\textit{``cat''} are the premise for understanding \textit{``golden dog''}+\textit{``white cat''}.
We investigate if existing vision-and-language models exhibit the consistency.
Our observations show that the models even with 37B parameters only achieve $\sim$40\% in the consistency, which indicates that existing models misunderstands the concepts in novel compositions.

In this paper, we propose a meta-learning based framework applicable to different types of models for consistent compositional generalization (CCG) across multiple levels of novel compositions.
The basic idea behind our framework is to progressively learn compositions from simple to complex by making models learn difficult compositions only after learning simple compositions.
To this end, we explicitly distinguish samples with different compositional complexities by generating different sample weights for them,
and adaptively update the sample weights to ensure learning compositions from simple to complex.
Specifically,
we divide the original training set into multiple validation sets based on their compositional complexities,
and introduce a set of meta-weight-nets to generate sample weights for samples in different validation sets.
To learn compositions from simple to complex, 
we make the model to fit the validation sets progressively in order of increasing compositional complexity, by training the model
and the meta-weight-nets independently and sequentially in a multilevel optimization manner.
In doing so, multiple levels of compositions are learned in a progressive consistent manner, thus achieving consistent compositional generalization across multiple levels of novel compositions.

To enable the quantitative evaluation for the consistency
of compositional generalization across multiple levels,
we build a new dataset in the context of visual question answering (VQA), \textit{i.e.}, GQA-CCG, based on the GQA dataset \cite{hudson2019gqa}, a large-scale dataset organized for compositional VQA.
We filter out samples that include novel compositions at the phrase-phrase level from the val\_all split of the GQA dataset to construct a sub-split.
We select some samples in the sub-split,
and manually annotate them with new questions containing novel compositions at simple levels including phrase-word level and word-word level for them.
Based on the annotated questions, we employ GPT-3.5 in an in-context learning manner to generate new questions for other samples in the sub-split, and conduct automatic postprocessing and manual review on the generated questions.
Furthermore, we introduce a consistency metric to measure whether a model achieves consistent compositional generalization across multiple levels.

We incorporate various types of methods of two tasks including VQA and temporal video grounding (TVG) into our framework, and conduct experiments on our GQA-CCG dataset, the GQA dataset and the Charades-CG dataset \cite{cvpr2022visa}, for validating the effectiveness and generalizability of our framework.
Experimental results show that
our framework effectively enhances the consistency of compositional generalization across multiple levels
and improves the accuracy of compositional generalization at different levels,
while maintaining comparable independent and identically distributed (IID) generalization capability.

To sum up, our contributions are as follows:
(1) To our knowledge, we are the first to explore the consistency of compositional generalization across multiple levels of novel compositions, which is critical for understanding the concepts in the novel compositions.
(2) We propose a meta-learning based framework for consistent compositional generalization across multiple levels of novel compositions.
(3) We present a GQA-CCG dataset to evaluate the consistency of compositional generalization across multiple levels of novel compositions for VQA models.

\section{Related work}
\subsection{Compositional Generalization}
Compositional generalization has received increasing attention as its importance in mimicking the fundamental compositionality of human cognition \cite{cognition2}.
Numerous benchmarks \cite{cvpr2022visa, li2024compositional} have been proposed to evaluate the compositional generalization capacity,
and a substantial amount of research \cite{neurips2020mgn, wang2023learning,li2023exploring} has been proposed to boost the compositional generalization capacity.

An important property regarding composition is that the process of composition is recursive \cite{bienenstock1996composition}, which revealed that 
compositions exist at multiple levels and 
compositional generalization capacity can be evaluated at multiple levels.
Recently, there have been several attempts that improve compositional generalization capacity at multiple levels.
For example,
works \cite{pierrot2019learning,liu2020compositional} perform recursive reasoning over a decomposed tree layout to achieve compositional generalization at multiple levels.
Yang \textit{et al.} \shortcite{yang2023deco} proposed a coarse-to-fine contrastive ranking loss for learning a composite representation that is sensitive to different levels of granularity of both queries and actions.
Xu \textit{et al.} \shortcite{xu2023meta} optimized models on multiple virtual sets in a bi-level optimization scheme to handle various levels of novel compositions.
These works focus on the accuracy on samples with multiple levels of novel compositions.
Differently,
we explore the consistency of compositional generalization across multiple levels, requiring a model to generalize not only to complex phrase-phrase level novel compositions but also to their associated simple phrase-word/word-word level novel compositions.

\subsection{Consistency}

Checking for consistency can be likened to conducting a Turing Test \cite{radziwill2017evaluating}, and the research community has demonstrated significant interest in assessing consistency.
Xu \textit{et al.} \shortcite{xu2018fooling} explored the consistency across image variations by performing adversarial attack on vision systems,
while \cite{shah2019cycle, ribeiro2019red} measure the consistency across linguistic variations by generating new questions with the same visual facts in the original question.
\cite{ray2019sunny, tascon2023logical} focus on logical consistency about logically consistent entailed questions or sufficient/necessary conditions.
\cite{selvaraju2020squinting, yuan2021perception} test perception consistency on low-level perception questions generated for reasoning questions.
Jing \textit{et al.} \shortcite{jing2022maintaining} improved reasoning consistency, which requires a VQA model make correct answers for a series of sub-questions about a compositional question.
Other works have also looked into consistency,
such as
spatial-temporal consistency \cite{wang2024self} in video-related tasks,
multi-view consistency \cite{yang2024viewfusion} and 2D-3D relational consistency \cite{zhang2024towards} in 3D-related tasks.
By contrast, we focus on the consistency of compositional generalization across multiple levels, which is critical for understanding novel compositions but remains unexplored.

\section{Framework}
\subsection{Overview}

\begin{figure}[t]
    \centering
    \includegraphics[width=1.0\linewidth]{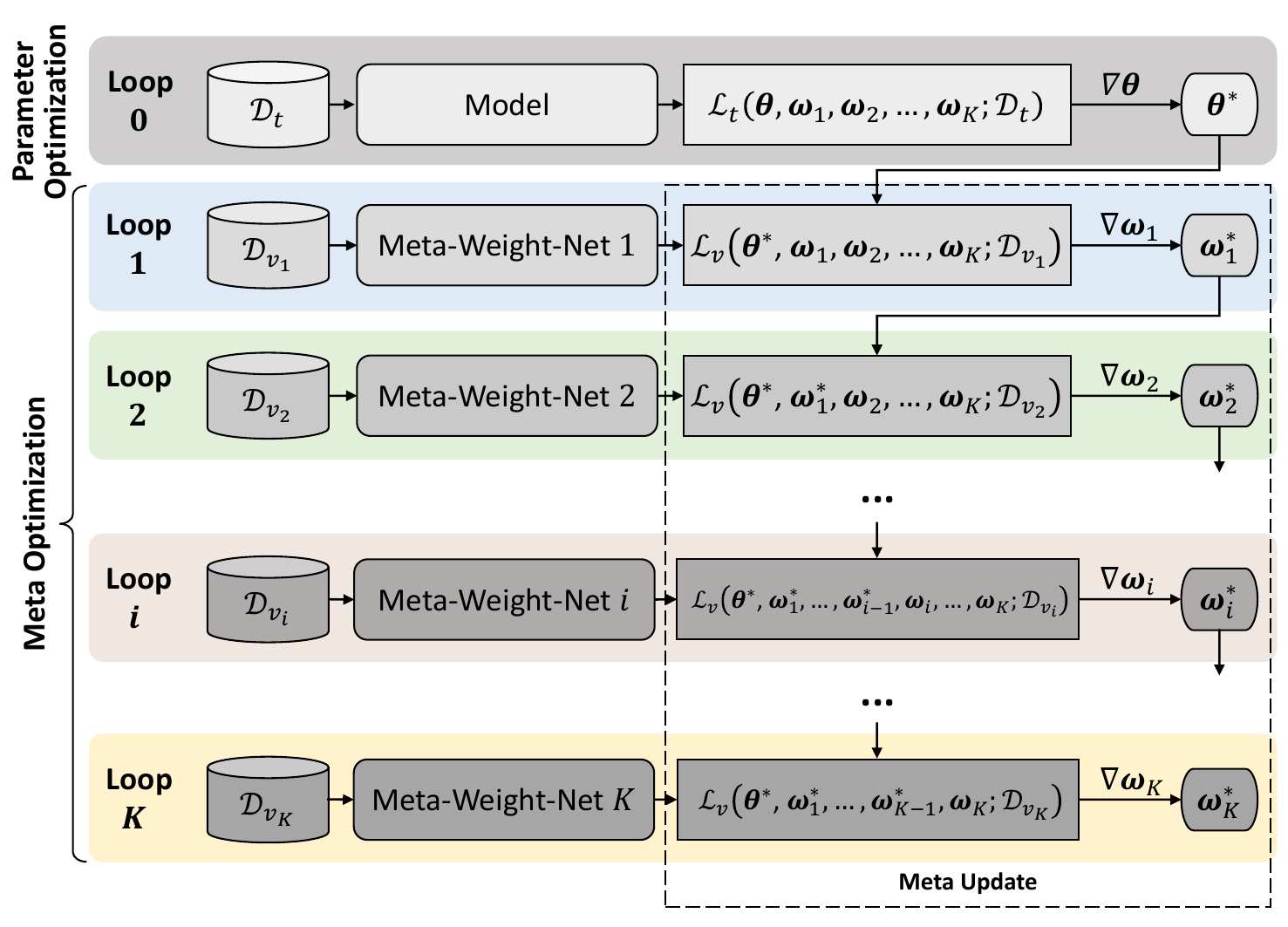}
	\caption{
             Overview of the proposed framework.
             }
    \label{fig:framework}
\end{figure}

The overview of the proposed framework is shown in Figure \ref{fig:framework}.
In our framework, we make models learn compositions from simple to complex by progressively fitting samples in order of increasing compositional complexity.
For a training set $\mathcal{D}_{t}$,
we first divide $\mathcal{D}_{t}$ into multiple validation sets $\{\mathcal{D}_{v_i}\}_{i=1}^{K}$ based on the compositional complexity of samples.
A larger $i$ indicates more complex samples in $\mathcal{D}_{v_i}$.
Then we introduce ${K}$ meta-weight-nets into the model, and use $i$-th meta-weight-net to generate sample weights for the samples in $\mathcal{D}_{v_i}$.
Finally, we train the model and the meta-weight-nets via a multilevel optimization process \cite{migdalas2013multilevel,choe2023betty},
where parameters of the model and different meta-weight-nets are optimized by their own unique objectives in a certain order.

To sum up,
we learn compositions in a progressive and consistent manner by:
(1) constructing validation sets to validate different training objectives of fitting samples with different complexities;
(2) using meta-weight-nets to generate sample weights that control which samples should be learned;
(3) updating the meta-weight-nets to fit the validation sets from simple to complex to control when to learn which samples.

\begin{figure}[!b]
    \centering
    \includegraphics[width=1.0\linewidth]{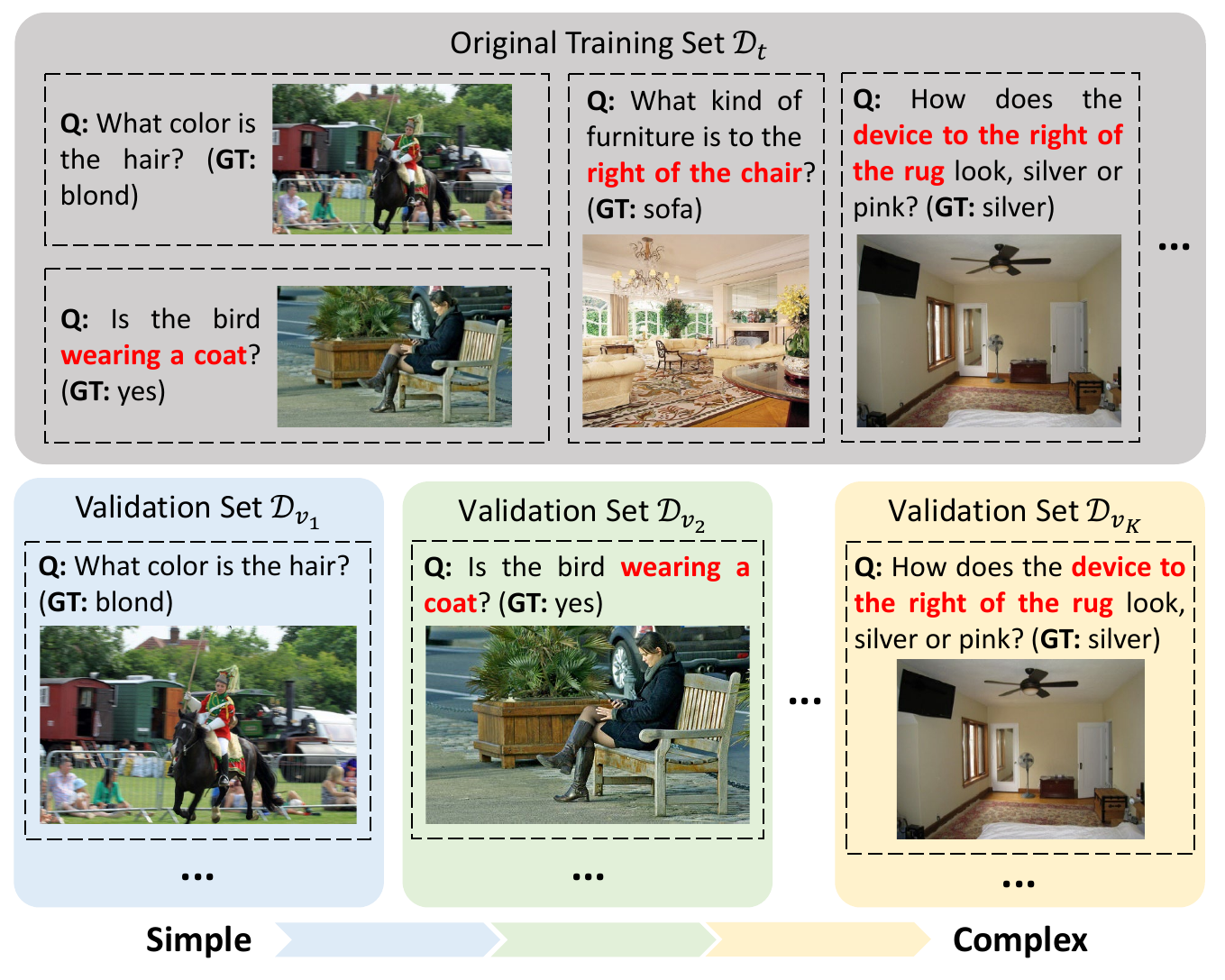}
	\caption{
             Validation set construction in the context of VQA, where the words in red font denote the longest phrase in the question.
             }
    \label{fig:validation}
\end{figure}

\subsection{Validation Set Construction}

To learn samples with different levels of compositional complexity,
we divide the original training set to construct multiple validation sets.
Each validation set is expected to contain samples with a certain level of compositional complexity.
For clarity, we introduce how to construct validation sets in the context of VQA, as shown in Figure \ref{fig:validation}.
VQA requires models to learn to provide a correct answer $A$ for a natural language question $Q$ about an image $V$.
For a training set $\mathcal{D}_{t}$ of VQA,
compositional generalization is the capacity of a VQA model trained on $\mathcal{D}_{t}$ to correctly answer questions with novel compositions composed of primitives (\textit{i.e.}, words and phrases) seen in $\mathcal{D}_{t}$.

Specifically, given a sample $(Q, V, A) \in \mathcal{D}_{t}$,
we first use the benepar toolkit \cite{kitaev-etal-2019-multilingual} to extract phrases in $Q$.
The phrases are denoted as $\{P_i\}_{i=1}^{N_P}$, where $N_p$ represent the number of phrases.
Generally, the compositional complexity is proportional to the length of the longest phrase in a question, with longer longest phrases indicating more complex questions.
As a result, we count the length of the longest phrase in the question as an approximated compositional complexity, and denote it as $\mathrm{L}(Q)$.
Next, we count the number of samples whose longest phrase length is $L(Q)$ using a function $S(\cdot)$, and denote the number as $S(L(Q))$.
Finally, we construct validation sets according to two principles:
(1) Samples with similar compositional complexity should be placed in the same validation set as much as possible.
(2) The number of samples in each validation set should be as close as possible.
Based on these two principles, we construct the $i$-th validation set by using
\begin{equation}
    \begin{aligned}
    & \ \ \ \ \ \ \ \ \ \ \ \ \ \mathcal{D}_{v_i} = \Big\{  (Q, V, A)  | (Q, V, A) \in \mathcal{D}_{t}, \\
    & \sum\nolimits_{1 \leq j \leq L(Q)} S(j) \geq \lfloor \mathrm{max}(|\mathcal{D}_{t}| \times (i-1) / K, 0) \rfloor, \\
    & \sum\nolimits_{1 \leq j \leq L(Q)} S(j) < \lfloor \mathrm{max}(|\mathcal{D}_{t}| \times i / K, |\mathcal{D}_{t}|) \rfloor \Big\},
    \end{aligned}
\end{equation}
where $|\cdot|$ is the sample number of the input dataset, and $K$ is a hyperparameter denoting the number of expected validation sets.
In doing so, 
each sample in the original training set is assigned to a unique validation set and satisfies
$\sum\nolimits_{1 \leq i \leq K} \mathcal{D}_{v_i} = \mathcal{D}_{t}$,
while samples in different validation sets have different complexities, 
the larger $i$ is, the more complex the samples in $\mathcal{D}_{v_i}$ are.

\subsection{Sample Weight Generation}

As different samples have different importance in learning a certain level of compositions,

we introduce a set of meta-weight-nets \cite{shu2019meta} to automatically generate sample weights for samples in different validation sets,
to explicitly control which samples should be learned.
Each meta-weight-net is only responsible for generating sample weights for a specific validation set.
These meta-weight-nets use the same architectures---three stacked fully connected layers and a sigmoid layer, but have different parameters.

For a sample $d \in \mathcal{D}_{v_i}$, 
$i$-th meta-weight-net is used to generate its sample weight.
The meta-weight-net accepts question features as input to output sample weight by
\begin{equation}
    w_d = f_i(\boldsymbol{\omega}_i; g(d)),
    \label{eq: sample weight}
\end{equation}
where $f_i(\cdot)$ and $\boldsymbol{\omega}_i$ denote the feedforward process and the parameters of the $i$-th meta-weight-net, respectively.
$g(\cdot)$ is the question feature of the input sample extracted by the language encoder of the VQA model.

\subsection{Multilevel Optimization}

To make the VQA model fit the validation sets progressively from simple to complex,
we optimize the model and the meta-weight-nets in a multilevel optimization process.
The process consists of two continuously alternating steps: parameter optimization and meta optimization.
During parameter optimization,
we freeze the meta-weight-nets, and optimize the model to fit current sample weights.
During meta optimization,
we update sample weights by optimizing meta-weight-nets to fit the validation sets.
The two steps are performed in sequence until the model training converges.
For model testing, we conventionally test the model and exclude the meta-weight-nets.
Below we first introduce the formulation of the multilevel optimization process, and then discuss the details of the parameter optimization and the meta 
optimization in the process.

\noindent
\textbf{Formulation.}
Let $\boldsymbol{\theta}$ and $\boldsymbol{\omega}_i$ denote the parameters of the model and the $i$-th meta-weight-net, respectively,
the multilevel optimization process can be formulated as sequentially performing the following nested loops from $LOOP_{0}$ to $LOOP_{K}$:
\begin{equation}
    \begin{aligned}
        & LOOP_{K}: \boldsymbol{\omega}^*_K = \mathop{\arg\min}\limits_{\boldsymbol{\omega}_K} \mathcal{L}_{v}(\boldsymbol{\theta}^*, \{ \boldsymbol{\omega}^*_{j} \}_{j=1}^{K-1}, \boldsymbol{\omega}_K; \mathcal{D}_{v_K}) \\
        & \ \ \ \ \ \ \ \ \ \ \ \ \ \ \ \ddots \\
        & LOOP_{2}: \ \ \mathrm{s.t.} \ \boldsymbol{\omega}^*_{2} = \mathop{\arg\min}\limits_{\boldsymbol{\omega}_{2}} \mathcal{L}_{v}(\boldsymbol{\theta}^*, \boldsymbol{\omega}^*_{1}, \boldsymbol{\omega}_2, ..., \boldsymbol{\omega}_K; \mathcal{D}_{v_2}) \\
        & LOOP_{1}: \ \ \ \ \ \ \ \mathrm{s.t.} \ \boldsymbol{\omega}^*_{1} = \mathop{\arg\min}\limits_{\boldsymbol{\omega}_{1}} \mathcal{L}_{v}(\boldsymbol{\theta}^*, \boldsymbol{\omega}_{1}, ..., \boldsymbol{\omega}_K; \mathcal{D}_{v_1}) \\
        & LOOP_{0}: \ \ \ \ \ \ \ \ \ \ \ \ \ \mathrm{s.t.} \ \boldsymbol{\theta}^* = \mathop{\arg\min}\limits_{\boldsymbol{\theta}} \mathcal{L}_{t}(\boldsymbol{\theta}, \boldsymbol{\omega}_{1}, ..., \boldsymbol{\omega}_K; \mathcal{D}_{t}),
    \end{aligned}
\end{equation}
where $\mathcal{L}_t$ and $\mathcal{L}_v$ denote the training loss and validation loss, respectively.
The $\mathcal{L}_t$ is determined by the selected model, as different models are trained by different losses.
Given a model trained by loss $\mathcal{L}$, by applying the proposed framework, $\mathcal{L}_t$ can be given by
\begin{equation}
    \mathcal{L}_t(\boldsymbol{\theta}, \boldsymbol{\omega}_1, \boldsymbol{\omega}_2, ..., \boldsymbol{\omega}_K; \mathcal{D}_{t}) = \sum\limits_{1 \leq i \leq K} \sum\limits_{d \in \mathcal{D}_{v_i}} w_d \mathcal{L}(\boldsymbol{\theta}; d),
\end{equation}
where $w_d$ is the sample weight of $d$ calculated by Eq.~\eqref{eq: sample weight}.
Furthermore, $\mathcal{L}_v$ can be written as
\begin{equation}
    \begin{aligned}
    \mathcal{L}_v(\boldsymbol{\theta}^*, \boldsymbol{\omega}^*_1, ..., \boldsymbol{\omega}^*_{i-1}, \boldsymbol{\omega}_{i}, ..., \boldsymbol{\omega}_K; \mathcal{D}_{v_i}) 
    = \sum\limits_{d \in \mathcal{D}_{v_i}} \mathcal{L}(\boldsymbol{\theta}^*(\omega_i); d).
    \end{aligned}
\end{equation}

\noindent
\textbf{Parameter Optimization.}
Parameter optimization aims to find the optimal parameters $\boldsymbol{\theta}^*$ such that minimizing the training loss $\mathcal{L}_t$.
Specifically, for the initial parameter $\boldsymbol{\theta}^{(0)}$, we train for $T_p$ iterations to update $\boldsymbol{\theta}$ by performing gradient descent.
At each iteration ($i=1, ..., T_p$), we update the parameter as follows:
\begin{equation}
    \boldsymbol{\theta}^{(i+1)} = \boldsymbol{\theta}^{(i)} - \beta_{\boldsymbol{\theta}}^{(i)} \nabla\boldsymbol{\theta}^{(i)},
\end{equation}
where $\nabla\boldsymbol{\theta}^{(i)} = \frac{d\mathcal{L}_t}{d\boldsymbol{\theta}^{(i)}}$, and $\beta_{\boldsymbol{\theta}}^{(i)}$ is the learning rate of $\boldsymbol{\theta}$ at iteration $i$.
We take the final parameter $\boldsymbol{\theta}^{(T_p)}$ as the optimal parameter $\boldsymbol{\theta}^{*}$.

\noindent
\textbf{Meta Optimization.}
At this step,
we find the optimal parameter $\boldsymbol{\omega}^*_i$ from $i=1$ to $K$ sequentially, for progressively fitting the validation sets from simple to complex.
For $i$-th meta-weight-net, we train it for $T_m$ iterations to update its parameter $\boldsymbol{\omega}_i$ to the optimal $\boldsymbol{\omega}^*_i$.
At each iteration ($j=1, ..., T_m$), we perform a meta update operation to $\boldsymbol{\omega}_i$ as follows:
\begin{equation}
    \boldsymbol{\omega}_i^{(j+1)} = \boldsymbol{\omega}_i^{(j)} - \beta_{\boldsymbol{\omega}_i}^{(j)} \nabla\boldsymbol{\omega}_i^{(j)},
\end{equation}
where $\nabla\boldsymbol{\omega}_i^{(j)} = \frac{d\mathcal{L}_v}{d\boldsymbol{\omega}_i^{(j)}}$, and $\beta_{\boldsymbol{\omega}_i}^{(j)}$ is the learning rate of $\boldsymbol{\omega}_i$ at iteration $j$.
However, $\nabla\boldsymbol{\omega}_i^{(j)}$ is difficult to directly calculate due to two aspects:
(1) There are multiple inevitable computationally onerous matrix-matrix multiplications during calculating $\nabla\boldsymbol{\omega}_i^{(j)}$.
(2) The calculation needs to save the calculation graph during multiple iterations, which increases the demand for GPU memory.
To solve this issue, we follow \cite{lorraine2020optimizing} with implicit function theorem to approximate the best-response Jacobian, and the details are provided in the \textbf{supplementary material}, which can be found at \url{https://github.com/NeverMoreLCH/CCG}.

\section{GQA-CCG Dataset}
\begin{figure}[t]
    \centering
    \includegraphics[width=1\linewidth]{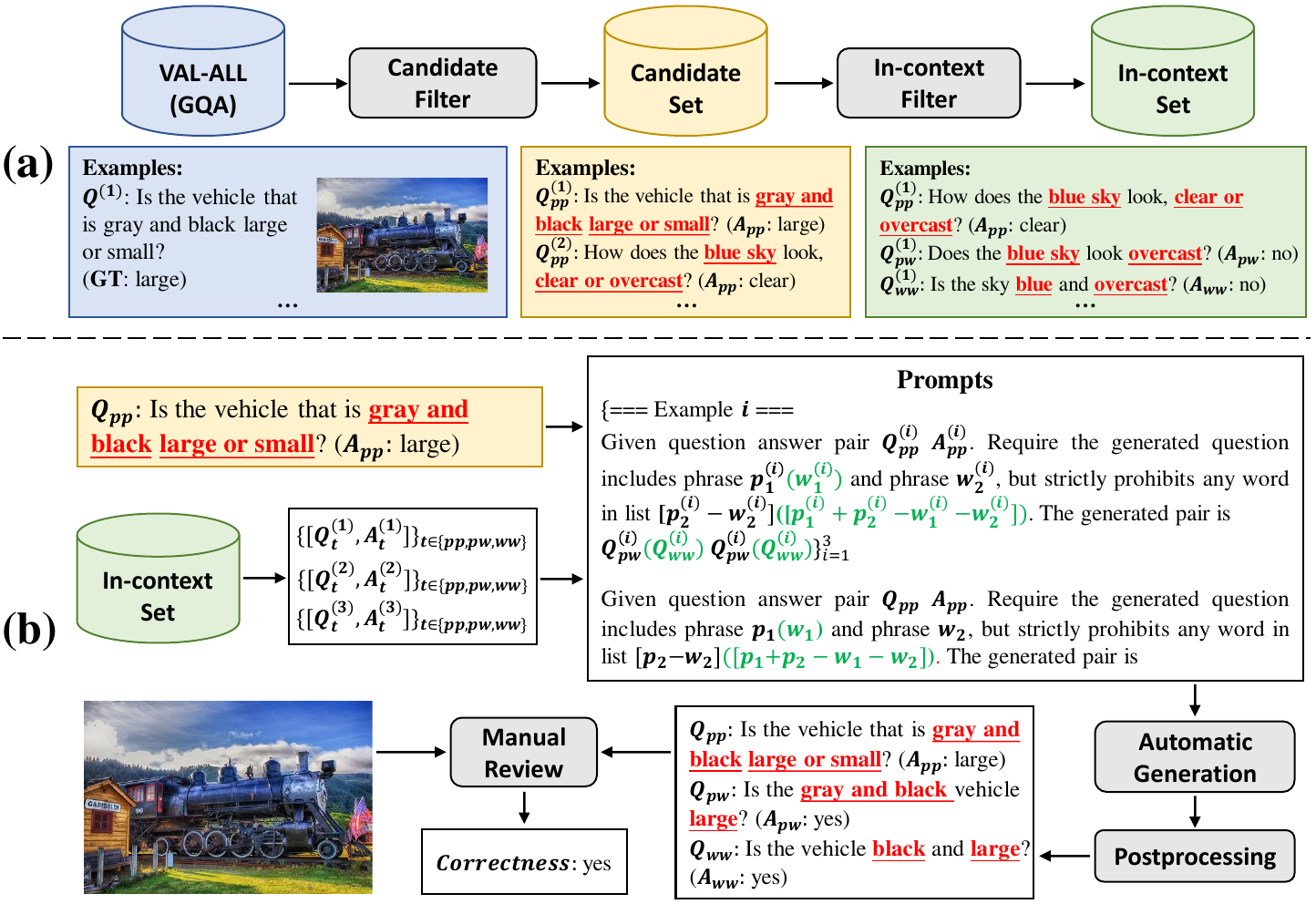}
	\caption{
             Overview of the pipeline for constructing GQA-CCG.
              (a) Preparations for constructing GQA-CCG.
			 (b) Illustration of the sample generation process.
             The underlined words/phrases in red font denote the components of novel compositions.
             The words in black font in prompts of (b) denote the prompt for generating $[Q_{pw}, A_{pw}]$.
             The words in green font in prompts of (b) denote the info that needs to be replaced when generating $[Q_{ww}, A_{ww}]$.}
    \label{fig:dataset}
\end{figure}

In this section, we illustrate how we construct GQA-CCG based on the GQA dataset, the overview of which is shown in Figure \ref{fig:dataset}.
We use the the train\_balanced split and the val\_all split of GQA in the process of constructing GQA-CCG, and here we denote them as $\mathcal{D}_t$ and $\mathcal{D}_v$, respectively.

\subsection{Preparations}

\textbf{Candidate Filter.}
First, we extract words and phrases by benepar \cite{kitaev-etal-2019-multilingual} for all questions in $\mathcal{D}_t$ and $\mathcal{D}_v$, respectively.
Then we count seen compositions including phrase-phrase (pp), phrase-word (pw) and word-word (ww) in $\mathcal{D}_t$.
Next, we filter out the samples in $\mathcal{D}_v$, whose question has at least a novel phrase-phrase composition.
We collected all question-answer pairs (denoted as $[Q_{pp}, A_{pp}]$) of the filtered samples as a candidate set $\mathcal{C}$.

\noindent
\textbf{In-context Filter.}
Based on $\mathcal{C}$, we select $M$ questions for each type of question prefix by a diversity maximization method.
For a type of question prefix,
we first randomly select a QA pair $[Q_{pp}, A_{pp}]$ of the type to construct an initial set $\mathcal{P}=\{[Q_{pp}, A_{pp}]\}$.
Then we find a QA pair, in which the question has the lowest average similarity to all questions in $\mathcal{P}$ of this type.
We add the found QA pair to $\mathcal{P}$, and repeat the above step until $|\mathcal{P}|=M$.
The similarities between the two questions are computed by the cosine similarity of their BERT embeddings \cite{devlin-etal-2019-bert}.
As a result, we obtain a set of in-context sets $\{\mathcal{I}_{t}\}_{t \in \mathcal{T}}$, where $\mathcal{T}$ denotes the set containing all types of question prefix,
and $\mathcal{I}_{t}$ represents the set of selected $M$ QA-pairs with type $t$.
Finally, we update $\mathcal{C} = \mathcal{C} - \mathcal{I}$, where $\mathcal{I} = \sum\nolimits_{t \in \mathcal{T}} \mathcal{I}_{t}$, and $\mathcal{I}_{t}$ can be represented as $\{[Q_{pp}^{(i)}, A_{pp}^{(i)}]\}_{i=1}^{M}$. 

For a QA pair $[Q_{pp}, A_{pp}] \in \mathcal{I}$, where $Q_{pp}$ has a novel phrase-phrase composition $p_1$-$p_2$ (notes that $p_1$ and $p_2$ can be exchanged), we manually annotate it with:
(1) A QA pair $[Q_{pw}, A_{pw}]$ with a novel phrase-word composition $p_1$-$w_2$.
(2) A QA pair $[Q_{ww}, A_{ww}]$ with a novel word-word composition $w_1$-$w_2$.
The relationship between $p_1$, $p_2$, $w_1$ and $w_2$ satisfies 
\begin{equation}
    w_1 \subsetneq p_1, \  w_2 \subsetneq p_2.
    \label{eq:sub}
\end{equation}
After the manual annotation, we rewrite $\mathcal{I}_{t}$ as $\{T_i\}_{i=1}^{M}$, where $T_i$ is a triplet that is denoted as $\{ [Q^{(i)}_t, A^{(i)}_t] \}_{t \in \{pp, pw, ww\}}$.

\subsection{Sample Generation Pipeline}

\textbf{Automatic Generation.}
For a QA pair $[Q_{pp}, A_{pp}] \in \mathcal{C}$ with type $t$, we first select 3 triplets, in which the questions have the maximum cosine similarity of BERT embeddings with $Q$ from $\mathcal{I}_{t}$.
We denote the novel phrase-phrase composition in $[Q_{pp}, A_{pp}]$ as $p_1$-$p_2$,
 $i$-th selected triplet as $P_i = \{[Q^{(i)}_{t}, A^{(i)}_{t}]\}_{t \in \{pp, pw, ww\}}$,
the components of novel compositions in $P_i$ as $p^{(i)}_1$, $p^{(i)}_2$, $w^{(i)}_1$ and $w^{(i)}_2$.
Then we iterate over the words in $p_2$ as $w_2$.
For each $w_2$, we iterate over the words in $p_1$ as $w_1$.
For each pair of $w_1$-$w_2$, we fill associated infos into the prompt template in Figure \ref{fig:dataset} (b),
and then use GPT-3.5 to generate $\{[Q_{t}, A_{t}]\}_{t \in \{pw, ww\}}$ to form a triplet $\{[Q_{t}, A_{t}]\}_{t \in \{pp, pw, ww\}}$.
Eventually, the generated triplets are collected as $\mathcal{G}$.

\noindent
\textbf{Postprocessing.}
For a triplet $\{[Q_t, A_t]\}_{t \in \{pp, pw, ww\}}$ in $\mathcal{G}$, we denote the novel phrase-phrase composition in $Q_{pp}$ as $p_1$-$p_2$, and retain the triplet that if it satisfies:
(1) There is a novel word-word composition $w_1$-$w_2$ and a novel phrase-word composition $p_1$-$w_2$ in $Q_{ww}$ and $Q_{pw}$, respectively.
(2) The relationship between $p_1$, $p_2$, $w_1$ and $w_2$ satisfies Eq.~\eqref{eq:sub}.
(3) There are no novel compositions that are not $w_1$-$w_2$ and $p_1$-$w_2$ in $Q_{ww}$ and $Q_{pw}$, respectively.

\noindent
\textbf{Manual Review.}
We recruited volunteers to verify the correctness of the generated QA pairs according to the image for the original QA pair.
For a triplet $\{ [Q_t, A_t] \}_{t \in \{pp, pw, ww\}}$ and the image $I$ for $[Q_{pw}, A_{pw}]$, we retain the triplet if it satisfies both $[Q_{pw}, A_{pw}]$ and $[Q_{ww}, A_{ww}]$ are correct based on $I$.
We add the associated images to the retained triplets, forming our dataset $\mathcal{D}_{CCG}$.
We get 18983 samples with novel compositions at different levels that consist of 8702 triplets of $\{ [Q_t, A_t, I] \}_{t \in \{pp, pw, ww\}}$.
The number of samples with novel compositions at phrase-phrase, phrase-word and word-word level are 5125, 8102 and 5756, respectively.

\subsection{Consistency Score}

To quantitatively evaluate the consistency on our $\mathcal{D}_{CCG}$, we devise a metric $Cons$, which is computed by
\begin{equation}
    \begin{aligned}
        Cons = \frac{\sum\nolimits_{T \in \mathcal{D}_{CCG}} \prod\nolimits_{t \in T} \mathrm{Correct}(t) }{\mathrm{triplet\_num}(\mathcal{D}_{CCG})},
    \end{aligned}
    \label{eq:consistency}
\end{equation}
where $\mathrm{Correct}(\cdot)$ is an indicator function, $\mathrm{triplet\_num}(\cdot)$ is a function that counts the triplet number of the input dataset.
The value range of $Cons$ is $[0, 1]$, and the larger $Cons$, the better the consistency.

\begin{table*}[t]
        \small
	\centering
        \begin{threeparttable}
	\setlength{\tabcolsep}{1.8mm}{
	\begin{tabular}{clccccc}
        \hline
		\multirow{2}{*}{Type} & \multirow{2}{*}{Method} & \multicolumn{4}{c}{Accuracy} & \multirow{2}{*}{Consistency} \\
        \cline{3-6}
        & & \textit{overall} & \textit{phrase-phrase} & \textit{phrase-word} & \textit{word-word} &   \\
        \hline
  	\multirow{2}{*}{Attention-based} & MAC \cite{iclr2018mac} & 62.07 & 70.97 & 59.84 & 57.28 & 30.82 \\
	& \textbf{+ MLO (Ours)} & 63.98 & 72.06 & 61.78 & 59.90 & 34.10 \\
        \hline
        \multirow{2}{*}{Graph-based} &  LCGN \cite{hu2019language} & 66.38 & 76.00 & 62.92 & 62.68 & 38.53 \\
        & \textbf{+ MLO (Ours)} & 67.53 & 76.92 & 64.14 & 63.86 & 40.46 \\
        \hline
        \multirow{2}{*}{NMN-based} &  MMN\cite{chen2021meta} & 70.14 & 84.87 & 66.57 & 62.07 & 42.41 \\
        & \textbf{+ MLO (Ours)} &  71.07 & \textbf{84.95} & 67.83 & 63.26 & 43.81 \\
        \hline
        \multirow{7}{*}{Pretrain-based} & OpenFlamingo (9B) \cite{awadalla2023openflamingo} & 53.47 & 54.15 & 49.67 & 58.20 & 19.58 \\ 
        \multirow{7}{*}{($\geq$ 7B)} & BLIP-2 ($\textup{FlanT5}_{\textup{XXL}}$) \cite{li2023blip} & 54.12 & 58.20 & 51.42 & 54.29 & 28.02 \\
        & Otter (7B) \cite{li2023otter} & 55.86 & 59.14 & 53.74 & 55.94 & 23.85 \\
        & LLaVA-1.5-Xtuner (7B) \cite{2023xtuner} & 65.28 & 57.66 & 66.14 & 70.85 & 35.89 \\
        & XComposer2 (7B) \cite{internlmxcomposer2} & 55.08 & 50.48 & 56.73 & 56.85 & 28.01 \\
        & mPLUG-Owl2 (7B) \cite{ye2023mplugowl2} & 65.08 & 58.22 & 65.96 & 69.96 & 36.32 \\
        & LLaVA-1.6 (7B) \cite{liu2024llavanext} & 61.50 & 59.06 & 61.91 & 63.10 & 34.52 \\
        & CogVLM (17B) \cite{wang2023cogvlm} & 67.47 &  61.37 &  68.56 &  71.37 & 38.30 \\
        & emu2 (37B) \cite{Emu2} & 68.46 & 63.22 & 68.65 & \textbf{72.86} & 40.58 \\
        \hline
        \multirow{4}{*}{Pretrain-based} & LXMERT \cite{tan2019lxmert} & 71.26 & 80.61 & 68.58 & 66.71 & 45.43 \\
        \cline{2-7}
        \multirow{4}{*}{($\leq$ 0.2B)} & VL-T5 \cite{cho2021unifying} & 70.19 & 77.22 & 67.66 & 67.51 & 42.78 \\
        & \textbf{+ MLO (Ours)} & 71.08 & 77.89 & 68.78 & 68.25 & 44.69 \\
        \cline{2-7}
        & CFR \cite{nguyen2022coarse} & 72.95 & 83.95 & 70.31 & 66.87 & 46.46 \\
        & \textbf{+ MLO (Ours)} & \textbf{74.23} & 84.50 & \textbf{71.81} & 68.75 & \textbf{49.27} \\
        \hline
	\end{tabular}}
	\end{threeparttable}
 \caption{Accuracy (\%) and Consistency (\%) of the state-of-the-art methods on GQA-CCG.}
\label{tab:gqa-ccg}
\end{table*}

\section{Experiment}

\subsection{Experimental Settings}

\noindent

\noindent
\textbf{Datasets.}
We apply the proposed framework to two tasks, VQA and TVG, to evaluate its effectiveness.
For VQA, we evaluate the framework on our GQA-CCG dataset and the GQA dataset \cite{hudson2019gqa}.
The GQA-CCG dataset is used to test the consistency of compositional generalization across multiple levels and the accuracy of compositional generalization at multiple levels. 
We use the GQA dataset to test whether our framework is harmful to the IID generalization capability.
For TVG, we use the recently released Charades-CG dataset \cite{cvpr2022visa} that contains compositional referring expressions about real-world videos to further test the compositional generalization capability of our framework on different tasks.
\noindent

\noindent
\textbf{Baseline Methods.}
For VQA, we incorporate our framework into five methods including MAC \cite{iclr2018mac}, LCGN \cite{hu2019language}, MMN \cite{chen2021meta}, VL-T5 \cite{cho2021unifying} and CFR \cite{nguyen2022coarse}.
and dub the incorporated methods $X$+MLO, where $X$ is a method name.
These methods belong to different types, 
thus the experiments on these methods allow for a comprehensive assessment of the effectiveness of our framework.
For TVG, we apply the proposed framework to MS-2D-TAN \cite{tpami2021ms2dtan},
which uses a 2D temporal map to model the temporal adjacent relations of video moments, and demonstrates good compositional capability.

\subsection{Compositional Generalization Performance}

We compare with different types of VQA methods including large vision-language models (LVLMs) varies in parameters (7B to 37B) on GQA-CCG, including MAC \cite{iclr2018mac}, LCGN \cite{hu2019language}, MMN \cite{chen2021meta}, OpenFlamingo \cite{awadalla2023openflamingo}, BLIP-2 \cite{li2023blip}, Otter \cite{li2023otter}, 
LLaVa-v1.5-Xtuner \cite{2023xtuner},
XComposer2 \cite{internlmxcomposer2},
mPLUG-Owl2 \cite{ye2023mplugowl2},
LLaVA-1.6 \cite{liu2024llavanext},
CogVLM \cite{wang2023cogvlm},
emu2 \cite{Emu2},
LXMERT \cite{tan2019lxmert},
VL-T5 \cite{cho2021unifying},
and CFR \cite{nguyen2022coarse}.
We evaluate pretrain-based models with parameters $\geq$ 7B via VLMEvalKit \cite{2023opencompass}, which provides model-specific prompts and answer matching rules.
We design additional matching patterns for each model with respect to its answer format.
For example, we use the matching pattern ``The answer is XXX." for XComposer2 as it often answers in this format.

The experimental results on GQA-CCG are listed in Table \ref{tab:gqa-ccg}, where ``\textit{overall}'' represents the accuracy on all test samples of GQA-CCG, and ``Consistency'' is the consistency score computed by Eq.~\eqref{eq:consistency}.
The ``\textit{phrase-phrase}'', ``\textit{phrase-word}'' and ``\textit{word-word}'' denote the accuracy on samples with corresponding levels of novel compositions.
We observe that:
(1) CFR+MLO achieves the best performance on both the accuracy and consistency (\textit{e.g.}, $74.23\%$ and $49.27\%$ in overall accuracy and consistency, respectively).
(2) For all five baseline methods of different types, our framework consistently improves their accuracy and consistency (\textit{e.g.}, $3.28\%$ and $2.81\%$  absolute performance gains in consistency for MAC and CFR, respectively).
(3) LVLMs are better at simple word-word level compositions than at complex phrase-phrase level and phrase-word level compositions.
Although several LVLMs outperform CFR+MLO in word-word accuracy,
they have more than thirty times the scale of parameters of CFR+MLO and have been trained on much more VQA samples.
These observations show that the proposed framework is efficient in improving not only the consistency but also the accuracy of compositional generalization at multiple levels for different types of baseline methods.
Furthermore, LVLMs still struggle to the consistency of compositional generalization, although they’ve been trained on a large amount of VQA samples.

\subsection{IID Generalization Performance}

The experimental results on GQA are listed in Table \ref{tab:gqa}.
We can observe that:
(1) Overall, CFR+MLO achieves the best performance among state-of-the-art methods.
(2) Compared to baseline methods, our framework improves their performance (\textit{e.g.}, $0.4\%$ absolute performance gains in accuracy for MAC).
The reason for the limited performance improvement of the proposed framework on GQA is that
we mainly focus on the compositional generalization capability, which can be viewed as a capability of out-of-distribution (OOD) generalization, while GQA is used more to evaluate the independent and identically distributed (IID) generalization.
The experimental results show that our framework is beneficial for IID setting (\textit{e.g.}, GQA) apart from the OOD setting (\textit{e.g.}, GQA-CCG), compared to most existing methods that provide performance gains in the OOD testing at the expense of IID performance \cite{cho2023generative}.

\begin{table}[tb]
	\small
	\centering
    \begin{threeparttable}[b]
	\setlength{\tabcolsep}{1.2mm}{
	\begin{tabular}{llc}
        \hline
		Type & Method & test-dev  \\
        \hline
        \multirow{2}{*}{Attention-based} & MAC \cite{iclr2018mac} & 52.43 \\
		                                 & \textbf{+ MLO (Ours)}  & 52.83 \\
        \cline{1-3}
        \multirow{2}{*}{Graph-based} & LCGN \cite{hu2019language} & 55.63 \\
        &  \textbf{+ MLO (Ours)} & 55.95 \\
        \cline{1-3}
        \multirow{2}{*}{NMN-based} & MMN \cite{chen2021meta} & 59.14 \\
                                   & \textbf{+ MLO (Ours)}  & 59.32 \\
        \cline{1-3}
        \multirow{1}{*}{Pretrain-based} & BLIP-2 ($\textup{FlanT5}_{\textup{XXL}}$) \cite{li2023blip} & 44.70 \\
        \multirow{1}{*}{(zero-shot)}& MiniGPT-4 \cite{zhu2023minigpt} & 43.50 \\
        \cline{1-3}
        \multirow{3}{*}{Pretrain-based} & VL-T5 \cite{cho2021unifying} & 58.40 \\
        \multirow{3}{*}{(fine-tuned)}&  \textbf{+ MLO (Ours)} & 58.59 \\
        \cline{2-3}
        & CFR \cite{nguyen2022coarse} & 70.27 \\
        &  \textbf{+ MLO (Ours)} & \textbf{70.51} \\
        \hline
	\end{tabular}}
    \end{threeparttable}
         \caption{Accuracy (\%) of the state-of-the-art methods on GQA \cite{hudson2019gqa}.}
         \label{tab:gqa}
\end{table}

\subsection{Ablation Studies}

Firstly, we investigate the effectiveness of different components of our framework on the consistency and accuracy of compositional generalization,
and the results are shown in Table \ref{tab:ablation1}.
We evaluate the effectiveness of meta-weight-nets by training them simultaneously rather than in a multilevel optimization manner.
We observe that the performance is better than the baseline methods but worse than the methods trained with our full framework,
which suggests that introducing meta-weight-nets is effective in improving the capability of compositional generalization and using multilevel optimization can obtain further improvements.
More ablation studies are provided in the \textbf{supplementary material},
including different variations of validation set construction, different optimization manners.

\begin{table}[tb]
	\small
	\centering
	\setlength{\tabcolsep}{1.2mm}{
	\begin{tabular}{cc|cc|cc}
        \hline
		\multirow{1.5}{*}{Meta-} & \multirow{1.5}{*}{Multilevel} & \multicolumn{2}{c|}{GQA-CCG} & \multicolumn{2}{c}{Charades-CG} \\
        \cline{3-4}\cline{5-6}
		Weight-Nets & Optimization & Acc & Cons & R1@0.5 & R5@0.5 \\
        \hline
		- & - & 62.07 & 30.82 & 42.04 & 77.16 \\
		$\checkmark$ & - & 63.12 & 32.54 & 43.73 & 78.39 \\
		$\checkmark$ & $\checkmark$ & \textbf{63.98} & \textbf{34.10} & \textbf{44.10} & \textbf{78.65} \\
        \hline
	\end{tabular}}
\caption{Ablation studies about components of our framework on GQA-CCG and the Novel-Composition split of Charades-CG \cite{cvpr2022visa}, on which we use MAC \cite{iclr2018mac} and MS-2D-TAN \cite{tpami2021ms2dtan} as baseline methods, respectively. The performance of baseline methods is shown in the first line.}
\label{tab:ablation1}
\end{table}

\subsection{Analysis of Validation Set Learning Sequence}

We analyze the importance of learning validation sets from simple to complex, to find whether learning from complex to simple is also effective to improve the compositional consistency.
The experimental results of using different learning procedures are list in Table \ref{tab:comp2simp}.
which shows that learning validation sets from complex to simple (C $\rightarrow$ S) has little improvement over the baseline model,
and even suffers a slight decline in some metrics (\textit{e.g.}, 0.14\% performance drop in phrase-phrase accuracy).
The main reason is that without the accumulation of simple knowledge, it is difficult to directly learn complex knowledge, which is proved in the human cognitive theory \cite{plass2010cognitive}.

\begin{table}[t]
	\small
	\centering
        \begin{threeparttable}
	\setlength{\tabcolsep}{0.6mm}{
	\begin{tabular}{cccccc}
		\hline
		\multirow{1.5}{*}{Learning} & \multicolumn{4}{c}{Accuracy} & \multirow{2}{*}{Cons} \\
        \cline{2-5}
        Sequence & \textit{overall} & \textit{phrase-phrase} & \textit{phrase-word} & \textit{word-word} &   \\
        \hline
  	- & 62.07 & 70.97 & 59.84 & 57.28 & 30.82 \\
        C $\rightarrow$ S & 62.05 & 70.83 & 59.89 & 57.28 & 30.95 \\
        S $\rightarrow$ C & \textbf{63.98} & \textbf{72.06} & \textbf{61.78} & \textbf{59.90} & \textbf{34.10} \\
        \hline
	\end{tabular}}
	\end{threeparttable}
 \caption{Performance of using different learning procedures of validation sets on GQA-CCG, where we use MAC \cite{iclr2018mac} as baseline methods (the first line), ``C $\rightarrow$ S" and ``S $\rightarrow$ C" denote learning from complex to simple and simple to complex, respectively.}
	\label{tab:comp2simp}
\end{table}

\subsection{Qualitative Analysis}

\begin{figure}[t]
    \centering
    \includegraphics[width=1\linewidth]{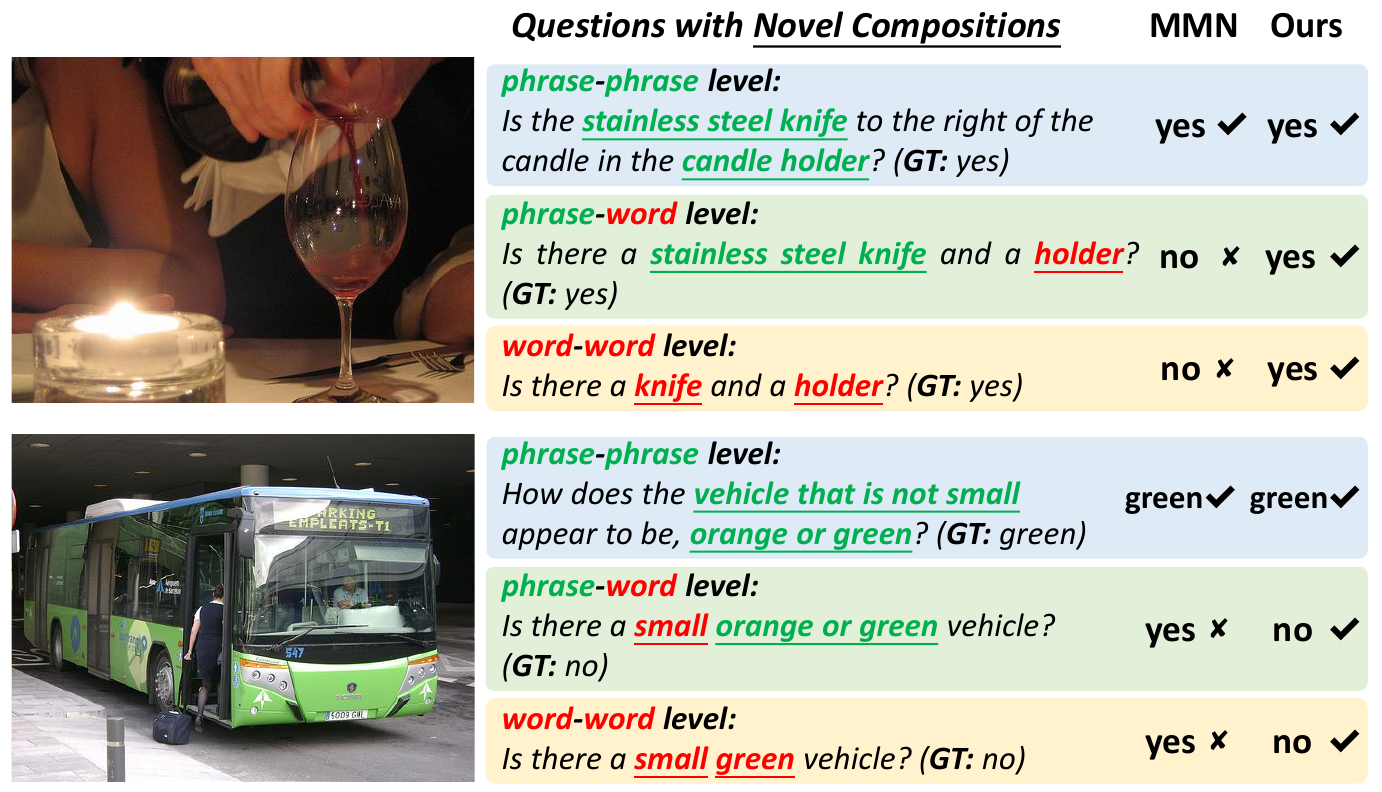}
	\caption{Qualitative comparisons between MMN+MLO (Ours) and MMN \cite{chen2021meta}.}
    \label{fig:qualitative}
\end{figure}

We provide several qualitative examples in the context of VQA in Figure \ref{fig:qualitative}.
For a triplet that consists of questions with novel compositions at different levels, we provide the predictions of MMN+MLO and MMN.
We can obverse that:
(1) MMN makes correct predictions for the questions with complex novel phrase-phrase compositions, but fails for the questions with associated simple phrase-word/word-word compositions.
(2) MMN+MLO (Ours) makes predictions accurately on all questions.
These observations show that our framework is effective to help the baseline method MMN maintain consistency of compositional generalization across different levels of novel compositions.
More qualitative examples are provided in the \textbf{supplementary material}.

\section{Conclusion}

In this paper, we have explored the consistency of compositional generalization across multiple levels of novel compositions,
and have presented that existing vision-and-language models even with 37B parameters struggle to the consistency.
We've proposed a meta-learning based framework that can improve the consistency of different models, by making the models progressively learn compositions from simple to complex in a multilevel optimization process.
Moreover, a GQA-CCG dataset has been presented to enable the qualitative evaluation of the consistency for VQA models.
Experimental results show that our framework 
can improve not only the consistency of compositional generalization across multiple levels,
but also the capacity of compositional generalization at different levels. 

\noindent
\textbf{Acknowledgments}
This work was supported by the Natural Science Foundation of China (NSFC) under Grants No. 62176021 and No. 62172041, Natural Science Foundation of Shenzhen under Grant No. JCYJ20230807142703006, Key Research Platforms and Projects of the Guangdong Provincial Department of Education under Grant No.2023ZDZX1034.

\bibliography{aaai25}

\begin{thebibliography}{50}
\providecommand{\natexlab}[1]{#1}

\bibitem[{Awadalla et~al.(2023)Awadalla, Gao, Gardner, Hessel, Hanafy, Zhu, Marathe, Bitton, Gadre, Sagawa, Jitsev, Kornblith, Koh, Ilharco, Wortsman, and Schmidt}]{awadalla2023openflamingo}
Awadalla, A.; Gao, I.; Gardner, J.; Hessel, J.; Hanafy, Y.; Zhu, W.; Marathe, K.; Bitton, Y.; Gadre, S.; Sagawa, S.; Jitsev, J.; Kornblith, S.; Koh, P.~W.; Ilharco, G.; Wortsman, M.; and Schmidt, L. 2023.
\newblock OpenFlamingo: An Open-Source Framework for Training Large Autoregressive Vision-Language Models.
\newblock \emph{arXiv preprint arXiv:2308.01390}.

\bibitem[{Bienenstock(1996)}]{bienenstock1996composition}
Bienenstock, E. 1996.
\newblock Composition.
\newblock In \emph{Brain theory}, 269--300. Elsevier.

\bibitem[{Chen et~al.(2021)Chen, Gan, Li, Cheng, Wang, and Liu}]{chen2021meta}
Chen, W.; Gan, Z.; Li, L.; Cheng, Y.; Wang, W.; and Liu, J. 2021.
\newblock Meta module network for compositional visual reasoning.
\newblock In \emph{Proceedings of the IEEE/CVF Winter Conference on Applications of Computer Vision}, 655--664.

\bibitem[{Cho et~al.(2021)Cho, Lei, Tan, and Bansal}]{cho2021unifying}
Cho, J.; Lei, J.; Tan, H.; and Bansal, M. 2021.
\newblock Unifying vision-and-language tasks via text generation.
\newblock In \emph{Proceedings of the International Conference on Machine Learning}, 1931--1942. PMLR.

\bibitem[{Cho et~al.(2023)Cho, Kim, Ryu, and Kweon}]{cho2023generative}
Cho, J.~W.; Kim, D.-J.; Ryu, H.; and Kweon, I.~S. 2023.
\newblock Generative bias for robust visual question answering.
\newblock In \emph{Proceedings of the IEEE/CVF Conference on Computer Vision and Pattern Recognition}, 11681--11690.

\bibitem[{Choe et~al.(2023)Choe, Neiswanger, Xie, and Xing}]{choe2023betty}
Choe, S.~K.; Neiswanger, W.; Xie, P.; and Xing, E. 2023.
\newblock Betty: An Automatic Differentiation Library for Multilevel Optimization.
\newblock In \emph{Proceedings of the International Conference on Learning Representations}.

\bibitem[{Contributors(2023{\natexlab{a}})}]{2023opencompass}
Contributors, O. 2023{\natexlab{a}}.
\newblock OpenCompass: A Universal Evaluation Platform for Foundation Models.
\newblock \url{https://github.com/open-compass/opencompass}.

\bibitem[{Contributors(2023{\natexlab{b}})}]{2023xtuner}
Contributors, X. 2023{\natexlab{b}}.
\newblock XTuner: A Toolkit for Efficiently Fine-tuning LLM.
\newblock \url{https://github.com/InternLM/xtuner}.

\bibitem[{Devlin et~al.(2019)Devlin, Chang, Lee, and Toutanova}]{devlin-etal-2019-bert}
Devlin, J.; Chang, M.-W.; Lee, K.; and Toutanova, K. 2019.
\newblock {BERT}: Pre-training of Deep Bidirectional Transformers for Language Understanding.
\newblock In \emph{Proceedings of the 2019 Conference of the North {A}merican Chapter of the Association for Computational Linguistics: Human Language Technologies, Volume 1 (Long and Short Papers)}, 4171--4186. Minneapolis, Minnesota: Association for Computational Linguistics.

\bibitem[{Dong et~al.(2024)Dong, Zhang, Zang, Cao, Wang, Ouyang, Wei, Zhang, Duan, Cao, Zhang, Li, Yan, Gao, Zhang, Li, Li, Chen, He, Zhang, Qiao, Lin, and Wang}]{internlmxcomposer2}
Dong, X.; Zhang, P.; Zang, Y.; Cao, Y.; Wang, B.; Ouyang, L.; Wei, X.; Zhang, S.; Duan, H.; Cao, M.; Zhang, W.; Li, Y.; Yan, H.; Gao, Y.; Zhang, X.; Li, W.; Li, J.; Chen, K.; He, C.; Zhang, X.; Qiao, Y.; Lin, D.; and Wang, J. 2024.
\newblock InternLM-XComposer2: Mastering Free-form Text-Image Composition and Comprehension in Vision-Language Large Model.
\newblock \emph{arXiv preprint arXiv:2401.16420}.

\bibitem[{Fodor and Pylyshyn(1988)}]{cognition2}
Fodor, J.~A.; and Pylyshyn, Z.~W. 1988.
\newblock Connectionism and cognitive architecture: A critical analysis.
\newblock \emph{Cognition}, 28(1-2): 3--71.

\bibitem[{Hu et~al.(2019)Hu, Rohrbach, Darrell, and Saenko}]{hu2019language}
Hu, R.; Rohrbach, A.; Darrell, T.; and Saenko, K. 2019.
\newblock Language-conditioned graph networks for relational reasoning.
\newblock In \emph{Proceedings of the IEEE/CVF International Conference on Computer Vision}, 10294--10303.

\bibitem[{Hudson and Manning(2018)}]{iclr2018mac}
Hudson, D.~A.; and Manning, C.~D. 2018.
\newblock Compositional Attention Networks for Machine Reasoning.
\newblock In \emph{Proceedings of the International Conference on Learning Representations}.

\bibitem[{Hudson and Manning(2019)}]{hudson2019gqa}
Hudson, D.~A.; and Manning, C.~D. 2019.
\newblock Gqa: A new dataset for real-world visual reasoning and compositional question answering.
\newblock In \emph{Proceedings of IEEE/CVF Conference on Computer Vision and Pattern Recognition}, 6700--6709.

\bibitem[{Jing et~al.(2022)Jing, Jia, Wu, Liu, and Wu}]{jing2022maintaining}
Jing, C.; Jia, Y.; Wu, Y.; Liu, X.; and Wu, Q. 2022.
\newblock Maintaining reasoning consistency in compositional visual question answering.
\newblock In \emph{Proceedings of IEEE/CVF Conference on Computer Vision and Pattern Recognition}, 5099--5108.

\bibitem[{Kitaev, Cao, and Klein(2019)}]{kitaev-etal-2019-multilingual}
Kitaev, N.; Cao, S.; and Klein, D. 2019.
\newblock Multilingual Constituency Parsing with Self-Attention and Pre-Training.
\newblock In \emph{Proceedings of the 57th Annual Meeting of the Association for Computational Linguistics}, 3499--3505. Florence, Italy: Association for Computational Linguistics.

\bibitem[{Li et~al.(2023{\natexlab{a}})Li, Zhang, Chen, Wang, Yang, and Liu}]{li2023otter}
Li, B.; Zhang, Y.; Chen, L.; Wang, J.; Yang, J.; and Liu, Z. 2023{\natexlab{a}}.
\newblock Otter: A Multi-Modal Model with In-Context Instruction Tuning.
\newblock \emph{arXiv preprint arXiv:2305.03726}.

\bibitem[{Li et~al.(2023{\natexlab{b}})Li, Li, Jing, Jia, and Wu}]{li2023exploring}
Li, C.; Li, Z.; Jing, C.; Jia, Y.; and Wu, Y. 2023{\natexlab{b}}.
\newblock Exploring the Effect of Primitives for Compositional Generalization in Vision-and-Language.
\newblock In \emph{Proceedings of IEEE/CVF Conference on Computer Vision and Pattern Recognition}, 19092--19101.

\bibitem[{Li et~al.(2024)Li, Li, Jing, Wu, Zhai, and Jia}]{li2024compositional}
Li, C.; Li, Z.; Jing, C.; Wu, Y.; Zhai, M.; and Jia, Y. 2024.
\newblock Compositional Substitutivity of Visual Reasoning for Visual Question Answering.
\newblock In \emph{Proceedings of the European Conference on Computer Vision}, 143--160. Springer.

\bibitem[{Li et~al.(2023{\natexlab{c}})Li, Li, Savarese, and Hoi}]{li2023blip}
Li, J.; Li, D.; Savarese, S.; and Hoi, S. 2023{\natexlab{c}}.
\newblock Blip-2: Bootstrapping language-image pre-training with frozen image encoders and large language models.
\newblock \emph{arXiv preprint arXiv:2301.12597}.

\bibitem[{Li et~al.(2022)Li, Xie, Qian, Zhu, Tang, Wu, Yang, Zhuang, and Wang}]{cvpr2022visa}
Li, J.; Xie, J.; Qian, L.; Zhu, L.; Tang, S.; Wu, F.; Yang, Y.; Zhuang, Y.; and Wang, X.~E. 2022.
\newblock Compositional temporal grounding with structured variational cross-graph correspondence learning.
\newblock In \emph{Proceedings of IEEE/CVF Conference on Computer Vision and Pattern Recognition}, 3032--3041.

\bibitem[{Liu et~al.(2024)Liu, Li, Li, Li, Zhang, Shen, and Lee}]{liu2024llavanext}
Liu, H.; Li, C.; Li, Y.; Li, B.; Zhang, Y.; Shen, S.; and Lee, Y.~J. 2024.
\newblock LLaVA-NeXT: Improved reasoning, OCR, and world knowledge.

\bibitem[{Liu et~al.(2020)Liu, An, Lou, Chen, Lin, Gao, Zhou, Zheng, and Zhang}]{liu2020compositional}
Liu, Q.; An, S.; Lou, J.-G.; Chen, B.; Lin, Z.; Gao, Y.; Zhou, B.; Zheng, N.; and Zhang, D. 2020.
\newblock Compositional generalization by learning analytical expressions.
\newblock \emph{Advances in Neural Information Processing Systems}, 33: 11416--11427.

\bibitem[{Lorraine, Vicol, and Duvenaud(2020)}]{lorraine2020optimizing}
Lorraine, J.; Vicol, P.; and Duvenaud, D. 2020.
\newblock Optimizing millions of hyperparameters by implicit differentiation.
\newblock In \emph{International conference on artificial intelligence and statistics}, 1540--1552. PMLR.

\bibitem[{Migdalas, Pardalos, and V{\"a}rbrand(2013)}]{migdalas2013multilevel}
Migdalas, A.; Pardalos, P.~M.; and V{\"a}rbrand, P. 2013.
\newblock \emph{Multilevel optimization: algorithms and applications}, volume~20.
\newblock Springer Science \& Business Media.

\bibitem[{Nguyen et~al.(2022)Nguyen, Do, Tran, Tjiputra, Tran, and Nguyen}]{nguyen2022coarse}
Nguyen, B.~X.; Do, T.; Tran, H.; Tjiputra, E.; Tran, Q.~D.; and Nguyen, A. 2022.
\newblock Coarse-to-Fine Reasoning for Visual Question Answering.
\newblock In \emph{Proceedings of IEEE/CVF Conference on Computer Vision and Pattern Recognition Workshops}, 4557--4565.

\bibitem[{Pierrot et~al.(2019)Pierrot, Ligner, Reed, Sigaud, Perrin, Laterre, Kas, Beguir, and de~Freitas}]{pierrot2019learning}
Pierrot, T.; Ligner, G.; Reed, S.~E.; Sigaud, O.; Perrin, N.; Laterre, A.; Kas, D.; Beguir, K.; and de~Freitas, N. 2019.
\newblock Learning compositional neural programs with recursive tree search and planning.
\newblock \emph{Advances in Neural Information Processing Systems}, 32.

\bibitem[{Plass, Moreno, and Br{\"u}nken(2010)}]{plass2010cognitive}
Plass, J.~L.; Moreno, R.; and Br{\"u}nken, R. 2010.
\newblock Cognitive load theory.

\bibitem[{Radziwill and Benton(2017)}]{radziwill2017evaluating}
Radziwill, N.~M.; and Benton, M.~C. 2017.
\newblock Evaluating quality of chatbots and intelligent conversational agents.
\newblock \emph{arXiv preprint arXiv:1704.04579}.

\bibitem[{Ray et~al.(2019)Ray, Sikka, Divakaran, Lee, and Burachas}]{ray2019sunny}
Ray, A.; Sikka, K.; Divakaran, A.; Lee, S.; and Burachas, G. 2019.
\newblock Sunny and Dark Outside?! Improving Answer Consistency in VQA through Entailed Question Generation.
\newblock In \emph{Proceedings of the Conference on Empirical Methods in Natural Language Processing and the International Joint Conference on Natural Language Processing}, 5860--5865.

\bibitem[{Ribeiro, Guestrin, and Singh(2019)}]{ribeiro2019red}
Ribeiro, M.~T.; Guestrin, C.; and Singh, S. 2019.
\newblock Are red roses red? evaluating consistency of question-answering models.
\newblock In \emph{Proceedings of the Annual Meeting of the Association for Computational Linguistics}, 6174--6184.

\bibitem[{Saqur and Narasimhan(2020)}]{neurips2020mgn}
Saqur, R.; and Narasimhan, K. 2020.
\newblock Multimodal graph networks for compositional generalization in visual question answering.
\newblock In \emph{Advances in Neural Information Processing Systems}, 3070--3081.

\bibitem[{Selvaraju et~al.(2020)Selvaraju, Tendulkar, Parikh, Horvitz, Ribeiro, Nushi, and Kamar}]{selvaraju2020squinting}
Selvaraju, R.~R.; Tendulkar, P.; Parikh, D.; Horvitz, E.; Ribeiro, M.~T.; Nushi, B.; and Kamar, E. 2020.
\newblock Squinting at vqa models: Introspecting vqa models with sub-questions.
\newblock In \emph{Proceedings of IEEE/CVF Conference on Computer Vision and Pattern Recognition}, 10003--10011.

\bibitem[{Shah et~al.(2019)Shah, Chen, Rohrbach, and Parikh}]{shah2019cycle}
Shah, M.; Chen, X.; Rohrbach, M.; and Parikh, D. 2019.
\newblock Cycle-consistency for robust visual question answering. In 2019 IEEE.
\newblock In \emph{Proceedings of IEEE/CVF Conference on Computer Vision and Pattern Recognition}, 6642--6651.

\bibitem[{Shu et~al.(2019)Shu, Xie, Yi, Zhao, Zhou, Xu, and Meng}]{shu2019meta}
Shu, J.; Xie, Q.; Yi, L.; Zhao, Q.; Zhou, S.; Xu, Z.; and Meng, D. 2019.
\newblock Meta-weight-net: Learning an explicit mapping for sample weighting.
\newblock \emph{Advances in neural information processing systems}, 32.

\bibitem[{Sun et~al.(2024)Sun, Cui, Zhang, Zhang, Yu, Wang, Rao, Liu, Huang, and Wang}]{Emu2}
Sun, Q.; Cui, Y.; Zhang, X.; Zhang, F.; Yu, Q.; Wang, Y.; Rao, Y.; Liu, J.; Huang, T.; and Wang, X. 2024.
\newblock Generative multimodal models are in-context learners.
\newblock In \emph{Proceedings of IEEE/CVF Conference on Computer Vision and Pattern Recognition}, 14398--14409.

\bibitem[{Tan and Bansal(2019)}]{tan2019lxmert}
Tan, H.; and Bansal, M. 2019.
\newblock LXMERT: Learning Cross-Modality Encoder Representations from Transformers.
\newblock In \emph{Proceedings of the Conference on Empirical Methods in Natural Language Processing and the International Joint Conference on Natural Language Processing}, 5100--5111.

\bibitem[{Tascon-Morales, M{\'a}rquez-Neila, and Sznitman(2023)}]{tascon2023logical}
Tascon-Morales, S.; M{\'a}rquez-Neila, P.; and Sznitman, R. 2023.
\newblock Logical Implications for Visual Question Answering Consistency.
\newblock In \emph{Proceedings of IEEE/CVF Conference on Computer Vision and Pattern Recognition}, 6725--6735.

\bibitem[{Wang et~al.(2024)Wang, Wu, Cen, Pan, Li, Wang, Cao, and Lin}]{wang2024self}
Wang, K.; Wu, Y.; Cen, J.; Pan, Z.; Li, X.; Wang, Z.; Cao, Z.; and Lin, G. 2024.
\newblock Self-Supervised Class-Agnostic Motion Prediction with Spatial and Temporal Consistency Regularizations.
\newblock In \emph{Proceedings of the IEEE/CVF Conference on Computer Vision and Pattern Recognition}, 14638--14647.

\bibitem[{Wang et~al.(2023{\natexlab{a}})Wang, Liu, Jing, Chen, Liang, Wang, and Shen}]{wang2023learning}
Wang, Q.; Liu, L.; Jing, C.; Chen, H.; Liang, G.; Wang, P.; and Shen, C. 2023{\natexlab{a}}.
\newblock Learning Conditional Attributes for Compositional Zero-Shot Learning.
\newblock In \emph{Proceedings of IEEE/CVF Conference on Computer Vision and Pattern Recognition}, 11197--11206.

\bibitem[{Wang et~al.(2023{\natexlab{b}})Wang, Lv, Yu, Hong, Qi, Wang, Ji, Yang, Zhao, Song et~al.}]{wang2023cogvlm}
Wang, W.; Lv, Q.; Yu, W.; Hong, W.; Qi, J.; Wang, Y.; Ji, J.; Yang, Z.; Zhao, L.; Song, X.; et~al. 2023{\natexlab{b}}.
\newblock Cogvlm: Visual expert for pretrained language models.
\newblock \emph{arXiv preprint arXiv:2311.03079}.

\bibitem[{Xu et~al.(2023)Xu, Huang, Shang, Yuan, Sun, and Liu}]{xu2023meta}
Xu, L.; Huang, M.~H.; Shang, X.; Yuan, Z.; Sun, Y.; and Liu, J. 2023.
\newblock Meta compositional referring expression segmentation.
\newblock In \emph{Proceedings of IEEE/CVF Conference on Computer Vision and Pattern Recognition}, 19478--19487.

\bibitem[{Xu et~al.(2018)Xu, Chen, Liu, Rohrbach, Darrell, and Song}]{xu2018fooling}
Xu, X.; Chen, X.; Liu, C.; Rohrbach, A.; Darrell, T.; and Song, D. 2018.
\newblock Fooling vision and language models despite localization and attention mechanism.
\newblock In \emph{Proceedings of the IEEE Conference on Computer Vision and Pattern Recognition}, 4951--4961.

\bibitem[{Yang et~al.(2023)Yang, Kong, Yang, Kehl, Sato, and Kobori}]{yang2023deco}
Yang, L.; Kong, Q.; Yang, H.-K.; Kehl, W.; Sato, Y.; and Kobori, N. 2023.
\newblock Deco: Decomposition and reconstruction for compositional temporal grounding via coarse-to-fine contrastive ranking.
\newblock In \emph{Proceedings of IEEE/CVF Conference on Computer Vision and Pattern Recognition}, 23130--23140.

\bibitem[{Yang et~al.(2024)Yang, Zuo, Ramasinghe, Bazzani, Avraham, and van~den Hengel}]{yang2024viewfusion}
Yang, X.; Zuo, Y.; Ramasinghe, S.; Bazzani, L.; Avraham, G.; and van~den Hengel, A. 2024.
\newblock ViewFusion: Towards Multi-View Consistency via Interpolated Denoising.
\newblock In \emph{Proceedings of the IEEE/CVF Conference on Computer Vision and Pattern Recognition}, 9870--9880.

\bibitem[{Ye et~al.(2024)Ye, Xu, Ye, Yan, Hu, Liu, Qian, Zhang, and Huang}]{ye2023mplugowl2}
Ye, Q.; Xu, H.; Ye, J.; Yan, M.; Hu, A.; Liu, H.; Qian, Q.; Zhang, J.; and Huang, F. 2024.
\newblock mplug-owl2: Revolutionizing multi-modal large language model with modality collaboration.
\newblock In \emph{Proceedings of IEEE/CVF Conference on Computer Vision and Pattern Recognition}, 13040--13051.

\bibitem[{Yuan et~al.(2021)Yuan, Wang, Jiang, and Chen}]{yuan2021perception}
Yuan, Y.; Wang, S.; Jiang, M.; and Chen, T.~Y. 2021.
\newblock Perception matters: Detecting perception failures of vqa models using metamorphic testing.
\newblock In \emph{Proceedings of IEEE/CVF Conference on Computer Vision and Pattern Recognition}, 16908--16917.

\bibitem[{Zhang et~al.(2021)Zhang, Peng, Fu, Lu, and Luo}]{tpami2021ms2dtan}
Zhang, S.; Peng, H.; Fu, J.; Lu, Y.; and Luo, J. 2021.
\newblock Multi-scale 2d temporal adjacency networks for moment localization with natural language.
\newblock \emph{IEEE Transactions on Pattern Analysis and Machine Intelligence}, 44(12): 9073--9087.

\bibitem[{Zhang, Luo, and Lei(2024)}]{zhang2024towards}
Zhang, Y.; Luo, H.; and Lei, Y. 2024.
\newblock Towards CLIP-driven Language-free 3D Visual Grounding via 2D-3D Relational Enhancement and Consistency.
\newblock In \emph{Proceedings of the IEEE/CVF Conference on Computer Vision and Pattern Recognition}, 13063--13072.

\bibitem[{Zhu et~al.(2023)Zhu, Chen, Shen, Li, and Elhoseiny}]{zhu2023minigpt}
Zhu, D.; Chen, J.; Shen, X.; Li, X.; and Elhoseiny, M. 2023.
\newblock Minigpt-4: Enhancing vision-language understanding with advanced large language models.
\newblock \emph{arXiv preprint arXiv:2304.10592}.

\end{thebibliography}

\end{document}